\documentclass[10pt,twocolumn,letterpaper]{article}

\usepackage[pagenumbers]{cvpr} % To force page numbers, e.g. for an arXiv version

\usepackage{graphicx}
\usepackage{amsmath}
\usepackage{amssymb}
\usepackage{booktabs}
\usepackage{float}

\usepackage[pagebackref,breaklinks,colorlinks]{hyperref}

\usepackage[capitalize]{cleveref}
\crefname{section}{Sec.}{Secs.}
\Crefname{section}{Section}{Sections}
\Crefname{table}{Table}{Tables}
\crefname{table}{Tab.}{Tabs.}

\def\cvprPaperID{CEE 247C Final Project} % *** Enter the CVPR Paper ID here
\def\confName{CVPR}
\def\confYear{2023}

\begin{document}

\title{Utility Pole Fire Risk Inspection from 2D Street-Side Images}

\author{Rajanie Prabha\\
Stanford University\\
{\tt\small rajanie@stanford.edu}
\and
Kopal Nihar\\
Stanford University\\
{\tt\small nkopal18@stanford.edu}
}
\maketitle

%%%%%%%%% ABSTRACT
\begin{abstract}
In recent years, California's electrical grid has confronted mounting challenges stemming from aging infrastructure and a landscape increasingly susceptible to wildfires. This paper presents a comprehensive framework utilizing computer vision techniques to address wildfire risk within the state's electrical grid, with a particular focus on vulnerable utility poles. These poles are susceptible to fire outbreaks or structural failure during extreme weather events. The proposed pipeline harnesses readily available Google Street View imagery to identify utility poles and assess their proximity to surrounding vegetation, as well as to determine any inclination angles. The early detection of potential risks associated with utility poles is pivotal for forestalling wildfire ignitions and informing strategic investments, such as undergrounding vulnerable poles and powerlines. Moreover, this study underscores the significance of data-driven decision-making in bolstering grid resilience, particularly concerning Public Safety Power Shutoffs. By fostering collaboration among utilities, policymakers, and researchers, this pipeline aims to solidify the electric grid's resilience and safeguard communities against the escalating threat of wildfires.
\end{abstract}

%%%%%%%%% BODY TEXT
\section{Introduction}
\label{sec:intro}
\begin{figure}[H]
\centering
  \includegraphics[height=7cm, width=7cm]{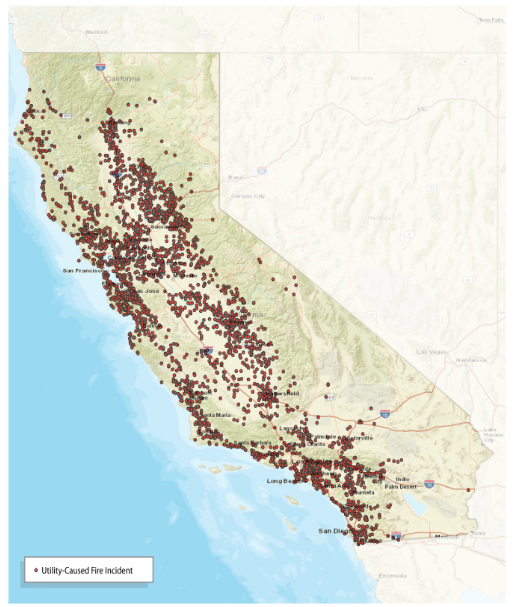}
    \caption{Locations of utility-caused ignitions in California}
    \label{fig:ignitions}

\end{figure}

According to US Forest Service, California's electrical grid 'is a sprawling network of aging power lines that overlaps with a drier landscape and more vulnerable to wildfires than ever. Dead trees and flammable foliage around grid components such as utility poles can fuel any ignitions induced by the electrical grid infrastructure. Between 2015 and 2020, Pacific Gas and Electric (PG\&E) reported 3550 utility-caused ignitions, also depicted in figure \ref{fig:ignitions}. The two main risk factors for utility poles are proximity with vegetation and pole inclination or tilt. The tilted poles can be highly vulnerable to falling during extreme weather events like wind storms. Many reports suggested leaning utility poles and bare electrical wires as the possible cause of the deadly Maui fires in Hawaii in 2023. Additionally, excessive vegetation around poles makes them highly risky. The 2021 Dixie Fire, a massive wildfire in Northern California, began when a tree fell onto a power line, causing an electrical fault, ultimately consuming 963,309 acres \cite{enwiki:1226601204}. 

To avoid such massive catastrophes, routine vegetation work around infrastructure is extremely important. Ideally, there are three zones: No trees in the wire zone, small trees or shrubs in the border zone, and trees no taller than the distance to the wire at maturity in the outer zone. 

\begin{figure}[H]
\centering
  \includegraphics[height=4cm, width=8cm]{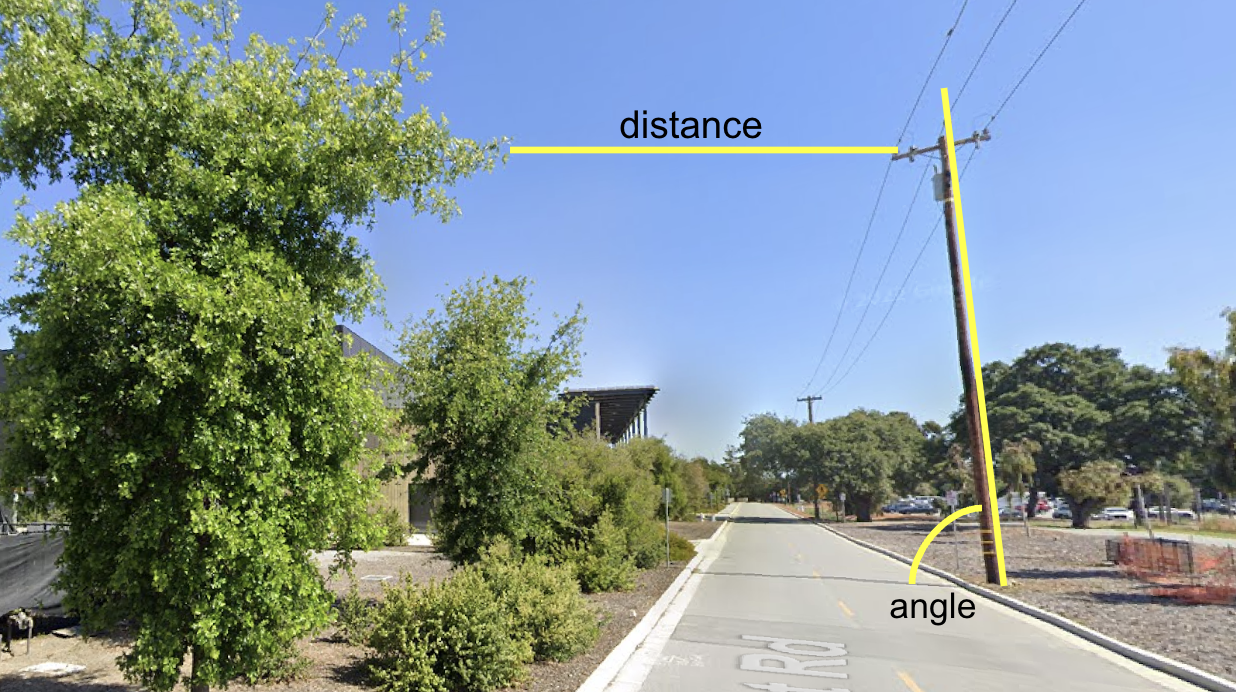}
    \caption{A Google Street View Image of a Utility Pole, showing two task objectives: distance from vegetation and pole inclination angle}
    \label{fig:objective}

\end{figure}

In this paper, we address the utility pole vulnerability by creating a pipeline that can assess the risk associated with the pole via two main factors: pole inclination and pole's proximity to vegetation. A Google Street View Image (GSV), along with the goals of our study are shown in figure \ref{fig:objective}. These two factors can be additionally combined with other pole characteristics such as age, material, etc, and other weather features such as wind speed to create more robust risk maps. Our pipeline leverages Google Street View imagery given their wide coverage and easy accessibility. For this study, San Francisco was selected as the focal area due to its heightened susceptibility to fire hazards and prevalent windy conditions.

We explore three different pipelines for these tasks. In the first approach, we collected 2D street-side images from GSV and located poles through a detection model. Following this, we employed a feature extraction technique to identify lines that closely match the detected poles. By analyzing the slope and intercept, we calculated the inclination angle of the pole. However, this approach is susceptible to variations in camera angles, which may introduce tilt to the poles, and it cannot assess vegetation proximity. In the second approach, we employed a depth estimation model to calculate the relative distance between the pole and the vegetation. Depth estimation models leverage deep learning techniques to infer the spatial arrangement and distance of objects within images or videos, facilitating the creation of depth maps that encode the scene's three-dimensional structure. These models analyze visual cues such as texture, perspective, and object size to predict depth information, enabling applications ranging from autonomous navigation to immersive virtual environments. However, this method does not provide us with information about the pole inclination. To establish a comprehensive approach capable of furnishing both metrics, we investigated an alternative pipeline centered on point clouds. Point clouds offer a nuanced portrayal of 3D surfaces by aggregating points in space, acquired through methods such as LiDAR \footnote{LiDAR (Light Detection and Ranging) is a remote sensing technique employing pulsed laser light to measure Earth's surface features and generate accurate three-dimensional representations of terrain.} or photogrammetry. Leveraging point clouds mitigates reliance on individual image camera perspectives and augments the feature space for enhanced risk analysis.

\section{Related Work}

Utility poles are highly vulnerable to natural disasters, particularly wildfires caused by tree contact with high-voltage power lines or tilted utility poles succumbing to pressure under high wind loads. Previous research has utilized LiDAR data to estimate distances between trees and power lines under varying high winds \cite{RemoteSensing}. Using pole segmentation and 3D reconstruction with point clouds, the authors analyzed the behavior of sample trees and validated against minimum vegetation clearing distances. Another study aimed to automatically calculate pole inclination angles by employing unmanned aerial vehicles for image collection and deep learning to detect and segment utility poles, with Hough transform to calculate angles based on segmented poles\cite{Zhu2019AutomaticUP}. Only a few studies have utilized Google Street View images for power line identification and vulnerability assessment using deep learning \cite{Springer1}. Our project builds on these efforts, advancing the field by leveraging easily accessible and cost-effective street view images to automatically measure utility pole inclination angles and detect their proximity to vegetation using point cloud reconstruction and depth estimation algorithms.

\section{Data}
\begin{figure}[H]
\centering
  \includegraphics[height=4.5cm, width=6cm]{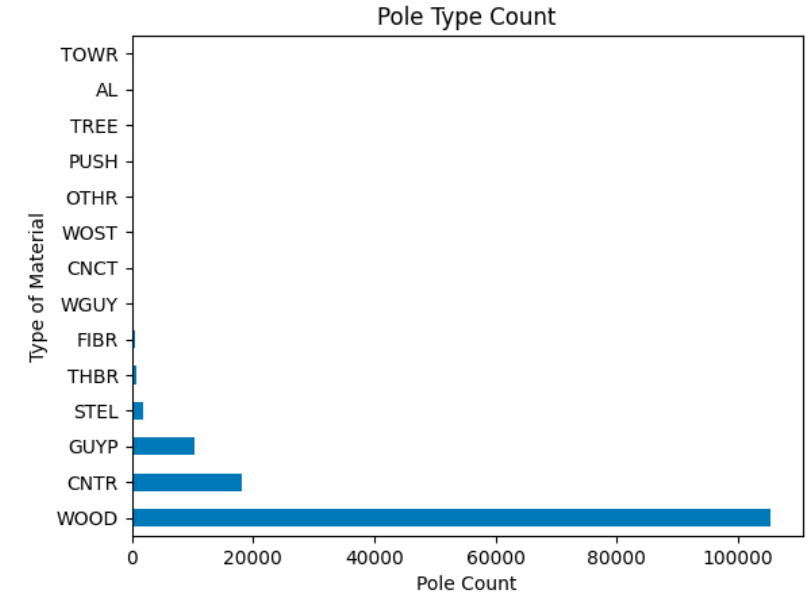}
    \caption{Count of pole types by material}
    \label{fig:poletype}
\end{figure}

Google Street View (GSV) imagery has become an invaluable asset across various academic disciplines, owing to its accessibility, extensive coverage, and high-resolution depiction of urban environments. Within our research pipeline, GSV serves as a primary resource for obtaining 2D street-level imagery, forming a crucial component of our data acquisition pipeline. Our data collection methodology involves a two-step process. Initially, we utilize the PG\&E Electric Distribution GIS Application data, which provides comprehensive information on grid components, including the precise geo-locations of utility poles and associated characteristics such as age, height, circumference, and material composition. According to the California state data, around 46\% of the poles are more than 50 years old and 76\% of the poles are made of wood. A histogram of pole types by material is shown in figure \ref{fig:poletype}.

\begin{figure}[H]
\centering
  \includegraphics[height=4.5cm, width=6cm]{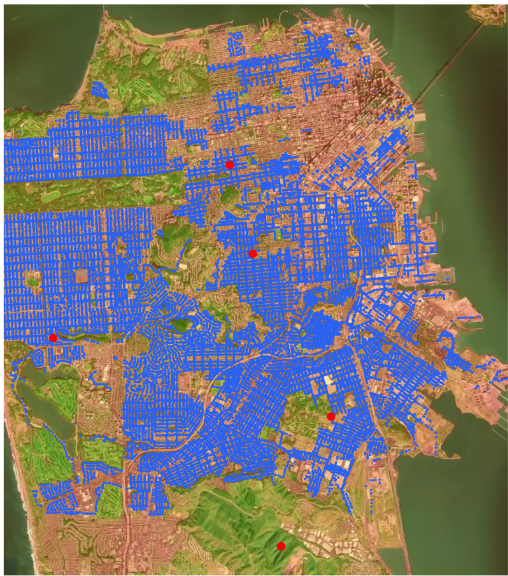}
    \caption{Distribution of vegetation (NVDI basemap), poles (blue), and ignitions (red) in SF}
    \label{fig:ndvi}
\end{figure}

Subsequently, rigorous data cleaning procedures were applied to refine the dataset, resulting in the identification and cataloging of approximately 2 million utility poles across California, as serviced by PG\&E. We narrow it down to only San Francisco geography. The next step is extracting street-view imagery around the pole and nearby vegetation for corresponding geo-locations, a distribution of which is shown in figure \ref{fig:ndvi}. The figure shows NDVI \footnote{Normalized difference vegetation index (NDVI) is a simple graphical indicator that is often used to analyze Remote Sensing measurements and assess whether the target being observed contains green health vegetation or not.} basemap with blue representing utility poles and red representing historically reported ignitions.

For the 2D imagery-based detection model, we use the maximum image size provided by GSV i.e. 620 by 620 with a field of view (fov) of 10, a pitch of 0, and a heading of 36 to capture 10 views around the geo-location. For the 3D point cloud reconstruction pipeline, we resize the imagery to 2500 by 2500 and extract additional images for different fov and pitch to provide more viewpoints for a better point cloud. 
% The images are of size 620 by 620, however, we resize them to (2500 by 2500) for better 3D point cloud reconstruction as the default size didn't perform well enough. For 2D The field of view selected is 60, with a pitch (the camera angle above the horizon) of 28 and 35 different angles for each utility pole.
\section{Methodology}

\subsection{Approach 1: Image Detection + Hough Transform}
\begin{figure*}[h]
  \centering
  \includegraphics[height=7cm,width=15cm]{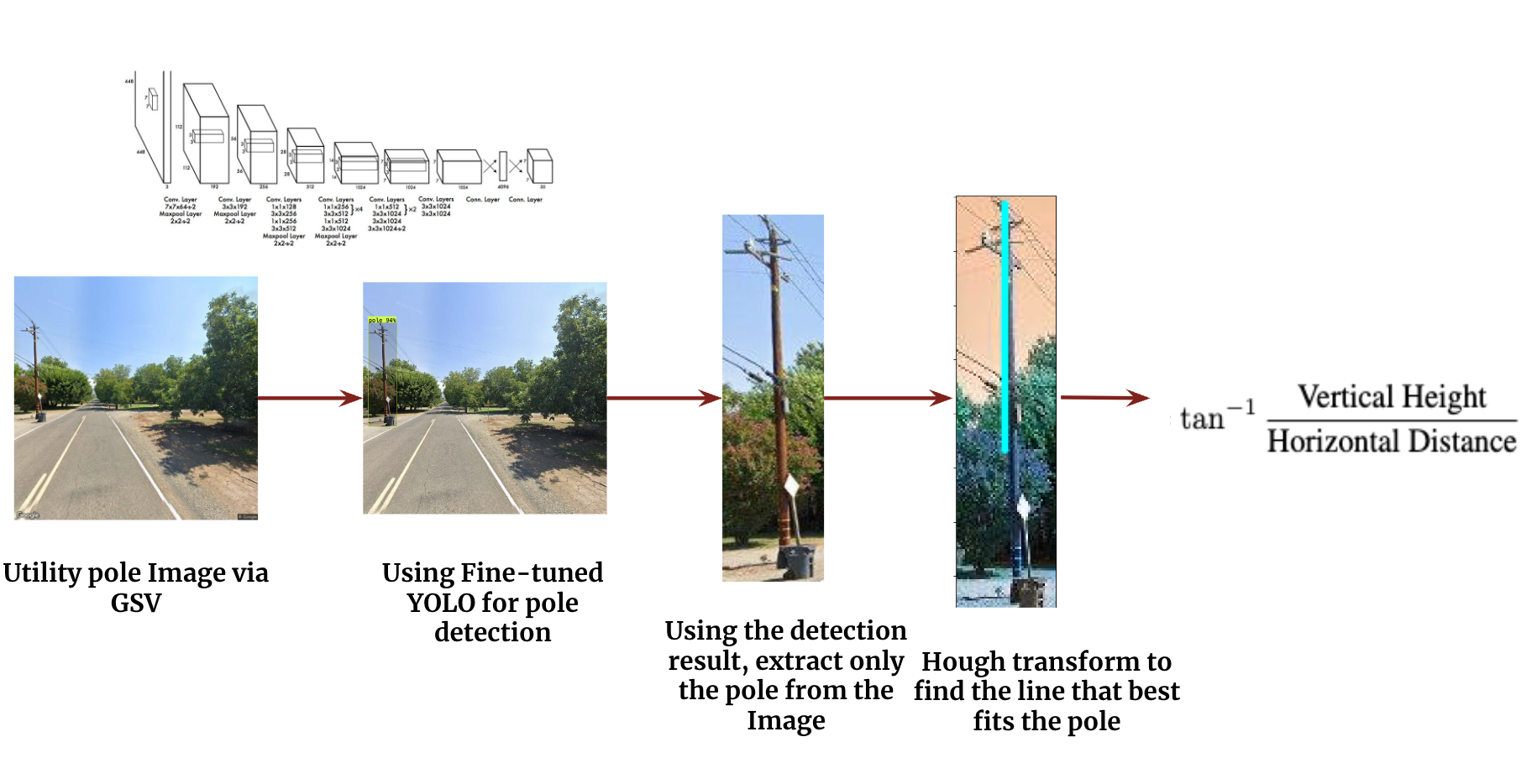}
    \caption{Methodology Pipeline for Image Detection + Hough Transform}
    \label{fig:2d-pipeline}

\end{figure*}

\begin{figure}[H]
\centering
  \includegraphics[height=4cm, width=8cm]{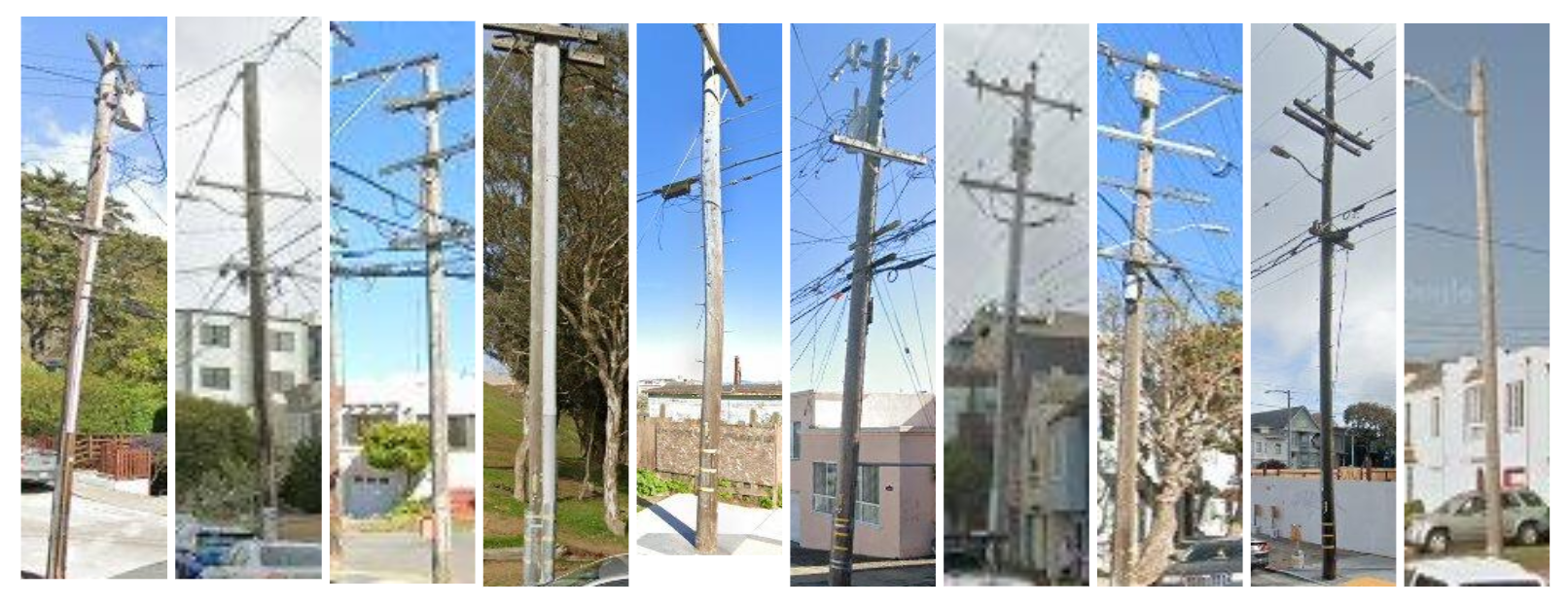}
    \caption{10 sample poles localized by YOLO detection model in inference mode}
    \label{fig:poles-detected}

\end{figure}

Inspired by the work of \cite{Zhu2019AutomaticUP}, we built a baseline pipeline for our task. As shown in \ref{fig:2d-pipeline}, we first extract the 2D street-side images from GSV. For each pole, 10 images are collected at the same fov and pitch but with different headings. We use You only look once (YOLO) detection model \cite{DBLP:journals/corr/RedmonDGF15} fine-tuned with a supervised training set of 1155 images and validated with 140 images. The test set has 15 images and the mean Average Precision (mAP) (defined in \cref{eval}) reported is 0.94. A few poles detected by the model are shown in figure \ref{fig:poles-detected}. 

Once we have localized the poles, hough transform \cite{10.1145/361237.361242} is applied to extract the lines that best fit the pole. The Hough transform works by representing the shapes as parameters in a mathematical space, typically known as the Hough space or parameter space. For detecting lines, each point in the image space corresponds to a line in the Hough space, defined by its slope and intercept. The algorithm then accumulates votes for potential lines by mapping the points in the image space to the corresponding lines in the Hough space. Finally, the peaks in the Hough space indicate the parameters of the lines present in the image. This method is robust against noise and variations in line orientation and widely used in applications such as edge detection, lane detection in autonomous vehicles, and shape recognition in computer vision tasks. Ultimately, the inclination angle can be determined by taking the arctangent of the vertical height divided by the horizontal distance. We call this the Inclination angle. The formula is represented as:
\begin{equation}
\text{Inclination angle} = \tan^{-1}\frac{\text{Vertical Height}}{\text{Horizontal Distance}}
\end{equation}

\begin{figure}[H]
\centering
  \includegraphics[height=4cm, width=8cm]{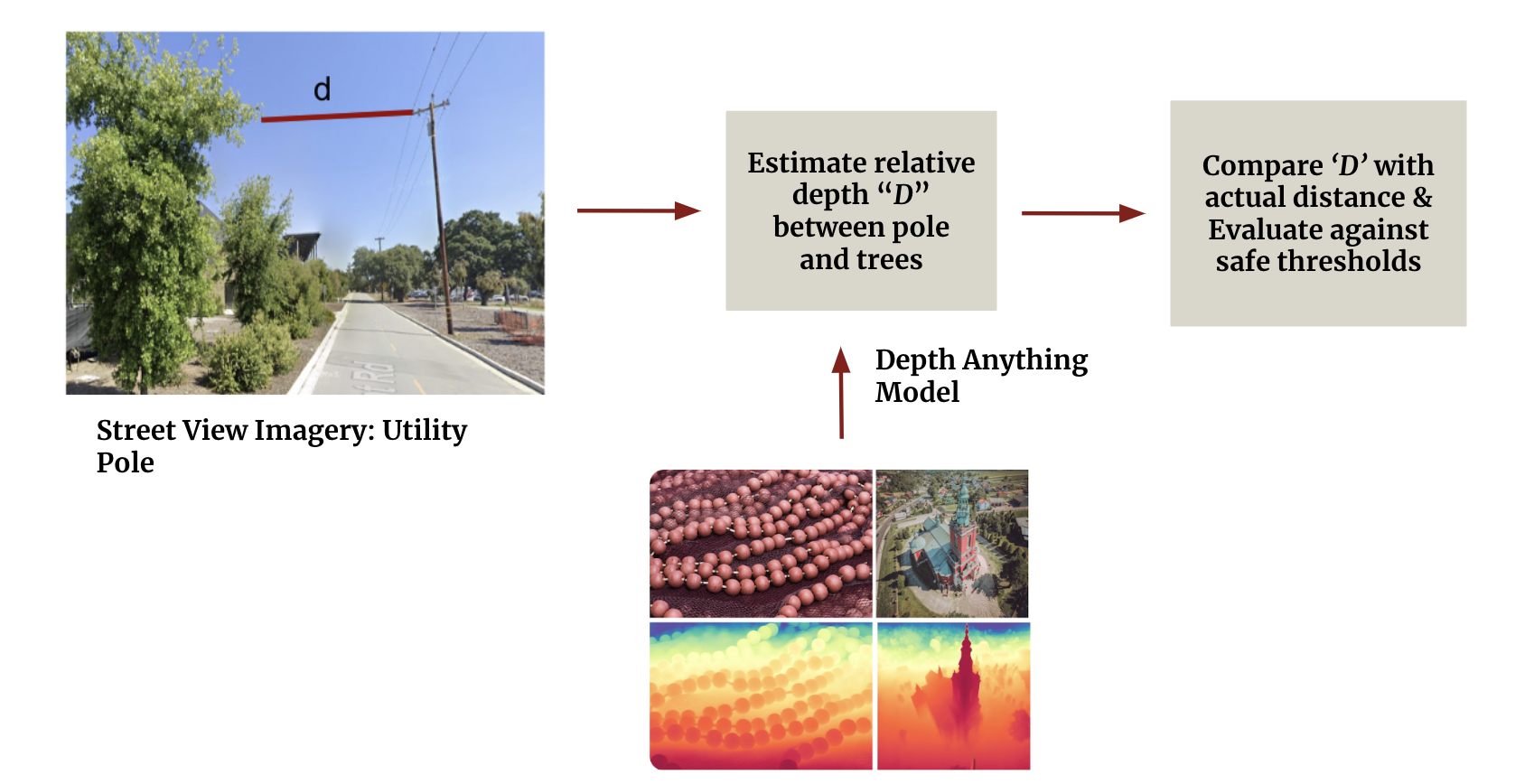}
    \caption{Methodology Pipeline for Depth Estimation}
    \label{fig:depth}
\end{figure}

\subsection{Approach 2: Depth Estimation}
To estimate the distance between vegetation and power lines, we employed the Depth Anything Model \cite{DepthAnything} to estimate the relative depth between them. Depth Anything, trained on a combination of 1.5M+ labeled images and 62M+ unlabeled images, is a highly efficient solution for robust monocular depth estimation. By carefully selecting street view images, we determined the relative depth between the utility poles and trees by subtracting the absolute depth value of the pole line from the absolute depth value of the trees. The pipeline is shown in figure \ref{fig:depth}.

We leveraged two pre-trained versions of the model - small and base.  Subsequently, we propose to compare and correlate the calculated relative depth with the actual distance obtained from Google Maps and evaluate it against predetermined safe thresholds. Although this method provides an underestimation of the actual distance between pole lines and trees, it serves as a valuable approximation for assessing the vulnerability of pole lines due to the proximity of vegetation.

\begin{figure}[H]
\centering
  \includegraphics[height=3.4cm, width=8cm]{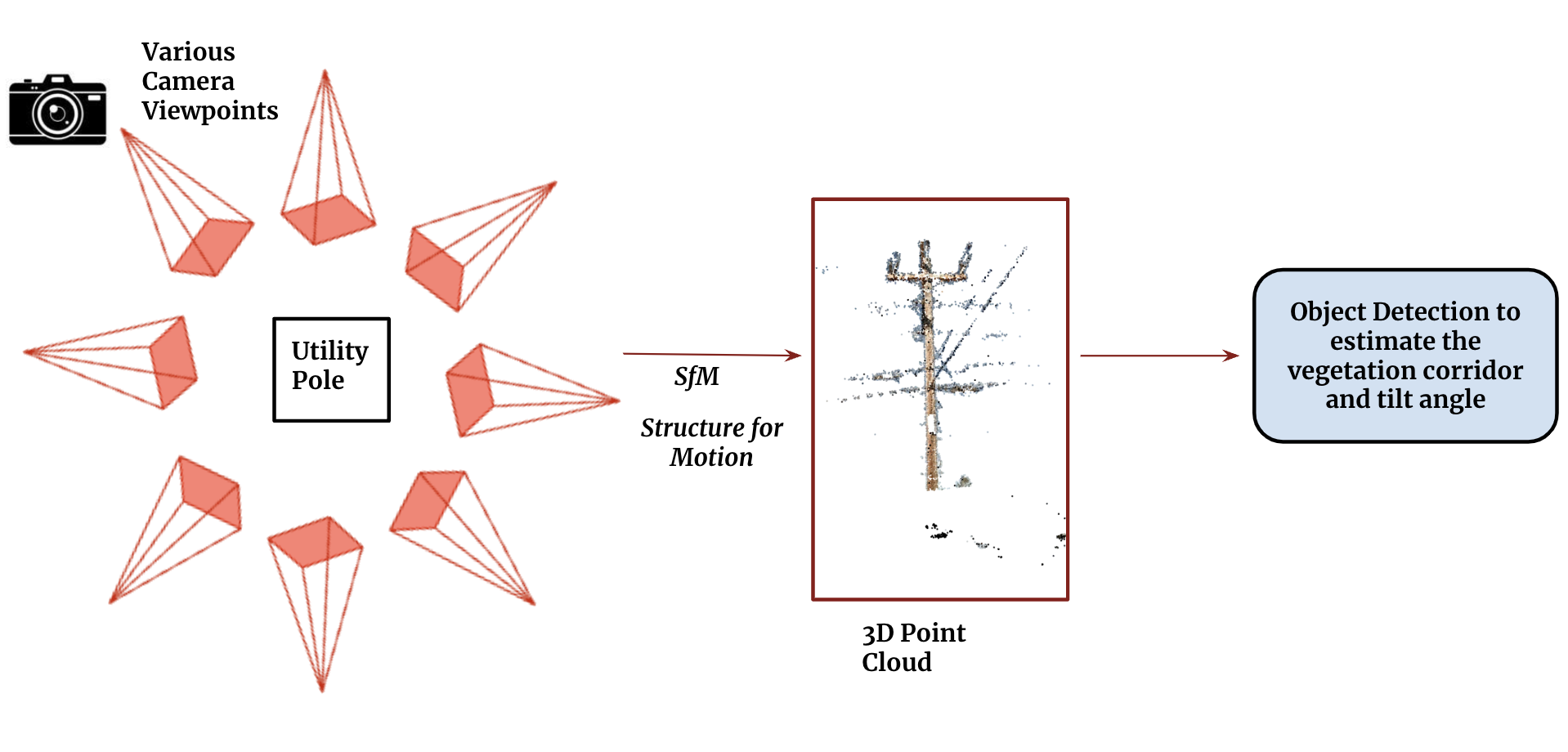}
    \caption{Methodology Pipeline for 3D point cloud reconstruction}
    \label{fig:sfm}
\end{figure}

\subsection{Approach 3: 3D Point Cloud Recontruction}
Estimating these variables from 2D images solely from 2D images may introduce challenges as it may require adjustments in the results due to variations in camera angles. Therefore, there is a need for a more comprehensive 3-dimensional analysis of the utility pole to estimate the vegetation proximity and inclination more accurately. This provides a more robust approach towards distorted views. Our pipeline is shown in figure \ref{fig:sfm}. First, we collect several images around the pole, capturing diverse viewpoints and resizing them for better reconstruction. For 3D point cloud reconstruction, we use the Structure for Motion (SfM) algorithm \cite{Schonberger_2016_CVPR}. SfM is a technique used to reconstruct three-dimensional scenes from two-dimensional images by analyzing the relative positions of visual features. It involves extracting key points from multiple images and computing their spatial relationships to estimate the shape and structure of the objects in the scene.

 For each utility pole, we leverage 35 2D street view images taken at various angles to create a 3D point cloud. In the next step, we use the Density-based spatial clustering of applications with noise (DBSCAN) algorithm \cite{10.5555/3001460.3001507} to segment the point cloud into various objects. DBSCAN works by iteratively examining each data point in the dataset and identifying its neighboring points within a specified distance threshold. If a point has a sufficient number of neighboring points within this threshold, it is classified as a core point. Core points that are close enough to each other are grouped into clusters. Points that do not meet the criteria to be core points but are within the neighborhood of a core point are classified as border points and are assigned to the same cluster as the core point. Data points that do not belong to any cluster and do not meet the density criteria are considered noise points. This process allows DBSCAN to automatically discover clusters of arbitrary shapes within the data, robustly handling outliers and noise. We have two main objects of interest (utility poles and vegetation). In the final step, we estimate the vegetation corridor and the tilt angle (via slope and intercept in the 3D plane). Once we have these factors, we can leverage additional data such as weather, and pole characteristics to create a fire-risk heatmap. 

\section{Evaluation Metrics}
\subsection{Mean Average Precision}\label{eval}
Mean Average Precision (mAP) in the context of the YOLO model represents the average precision across all classes for object detection tasks. It considers both precision and recall, evaluating the accuracy of the model's predictions by comparing them with ground truth annotations.

The formula for mAP is represented as:

\begin{equation}
\text{mAP} = \frac{1}{N} \sum_{i=1}^{N} \text{AP}_i
\end{equation}

where \(N\) is the number of classes and \(\text{AP}_i\) denotes the Average Precision for each class \(i\).

\subsection{Depth Anything Accuracy}
 While Google Maps provides absolute distances between the exact coordinates of poles and vegetation, the Depth Anything model estimates relative distances based on the distance of the image captured from the camera. A mathematical depiction is as follows: Let \textit{D} be the depth estimates from the Depth Anything Model, \textit{A} be the actual distances between poles and vegetation, and \textit{T} be the safe distance threshold. The accuracy metric  \(Acc\)  can be defined as the proportion of depth estimates that fall within the safe distance threshold. 

\begin{equation}
\small
    Acc = \frac{\text{Number of depth estimates within safe distance threshold}}{\text{Total number of depth estimates}}
\end{equation}

\subsection{Impact Metric}
To evaluate fire risk based on the estimated depth accuracy metric and safe distance thresholds, we define a fire risk score as follows:
\begin{equation}
\small
\text{Fire Risk} =
\begin{cases}
\text{Low}, & \text{if Accuracy} > \text{Thresh}_{\text{low}} \\
\text{Moderate}, & \text{if } \text{Thresh}_{\text{low}} \geq \text{Accuracy} > \text{Thresh}_{\text{mod}} \\
\text{High}, & \text{if Accuracy} \leq \text{Thresh}_{\text{mod}}
\end{cases} 
\end{equation}
Where:
Thresh(low) is the lower threshold for acceptable accuracy, indicating low fire risk. and Thresh(mod) is the upper threshold for acceptable accuracy, indicating moderate fire risk.

To evaluate the inclination angle risk, we propose a fragility metric for the utility pole, considering tilt angle, pole characteristics such as age and material, as well as wind speed, could be represented as:

\begin{equation}
    \text{Fragility} = f(\text{angle}, \text{pole features}, \text{wind})
\end{equation}

where \(f\) is a function that combines the effects of tilt, age, material, and wind speed. We define the Topple risk as follows:
\begin{equation}
\small
\text{Topple Risk} =
\begin{cases}
\text{Low}, & \text{if Fragility} > \text{Thresh}_{\text{low}} \\
\text{Moderate}, & \text{if } \text{Thresh}_{\text{low}} \geq \text{Fragility} > \text{Thresh}_{\text{mod}} \\
\text{High}, & \text{if Fragility} \leq \text{Thresh}_{\text{mod}}
\end{cases} 
\end{equation}

\subsection{Cost Metric}
In addition to the evaluation metrics for inclination angles and proximity of poles to vegetation, we also consider the following cost metric that provides insights into the cost-effectiveness of our proposed system in mitigating wildfire risks:
\begin{equation}
\small
    \text{Cost Metric} = \frac{\text{Implementation Costs}}{\text{Fire Risk}}
\end{equation}

Here, Implementation Costs represent the total expenses incurred in deploying the computer vision system for assessing fire risk and Fire Risk is a quantitative measure indicating the likelihood of a wildfire occurring due to factors evaluated in the study. A lower cost metric suggests that the system offers greater value in reducing fire risk relative to the investment required for implementation.

\section{Results}
\subsection{Image Detection + Hough Transform}
\begin{figure*}[htbp]
    \centering
    \includegraphics[width=0.4\textwidth]{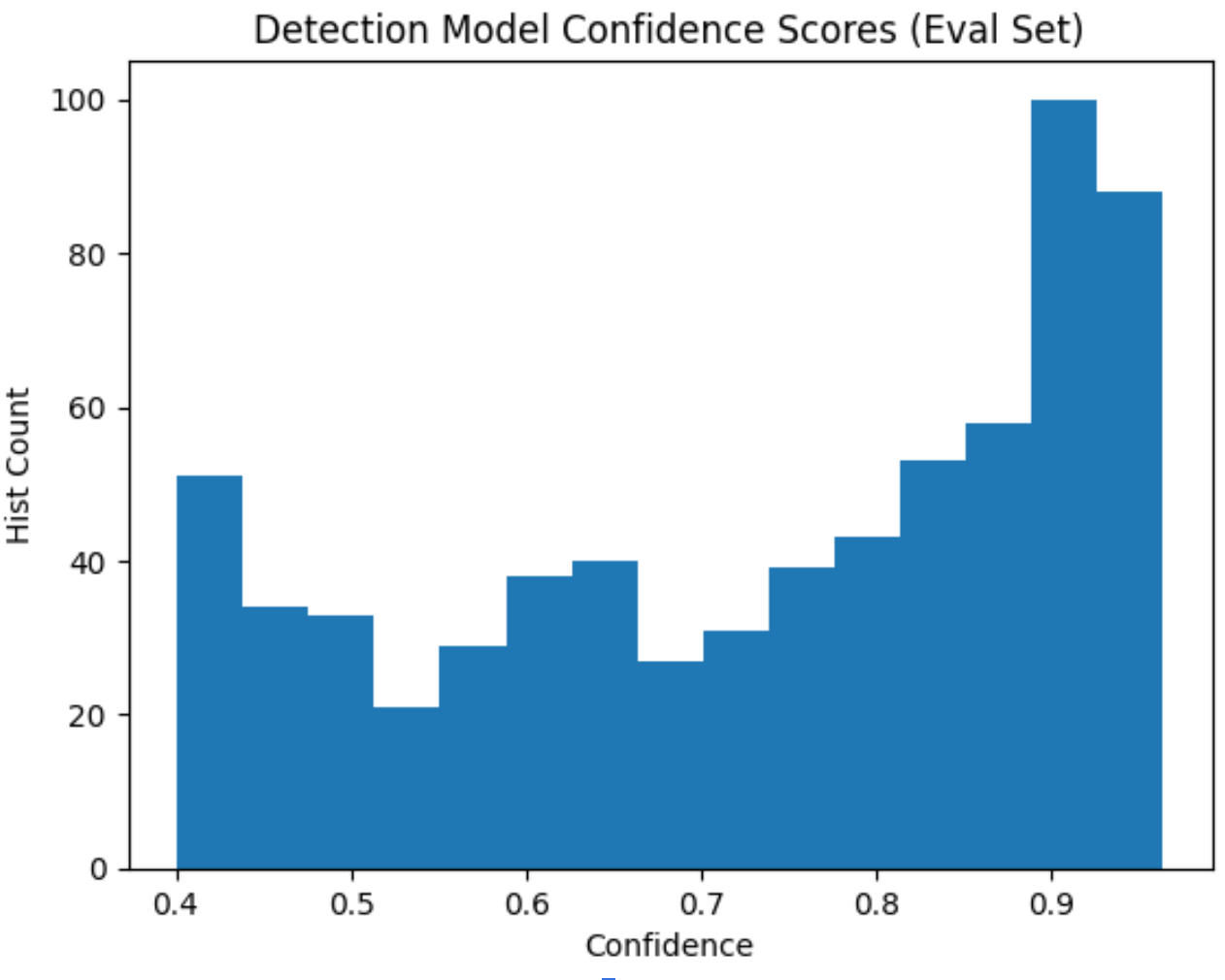}
    \hfill
    \includegraphics[width=0.4\textwidth]{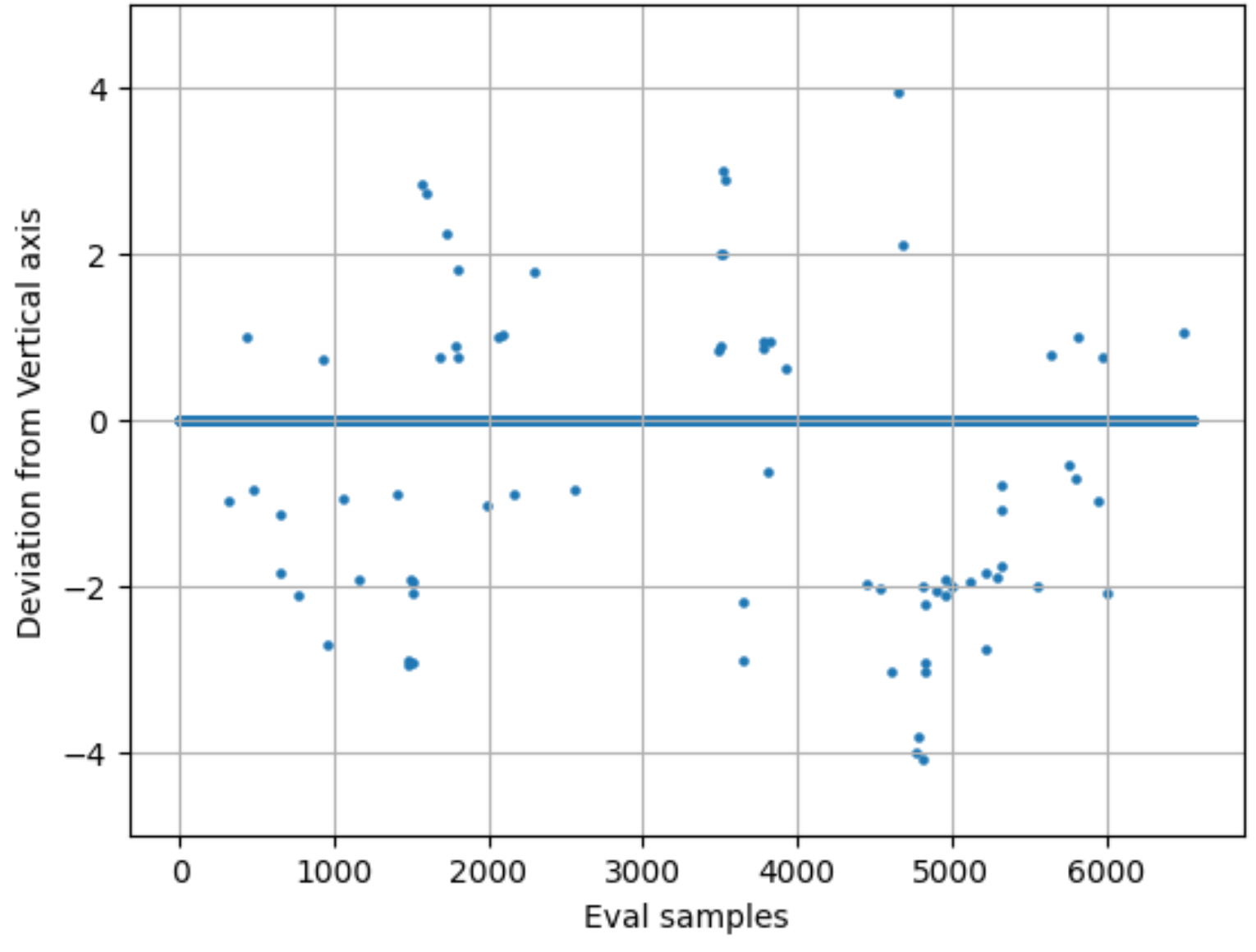}
    \caption{(left) Confidence scores of the detection results on 100 poles obtained from the YOLO model, (right) Final result showing pole deflection from the vertical axis (in degrees).}
    \label{fig:results1}
\end{figure*}

We evaluated our first pipeline on 100 randomly selected poles collected across San Francisco County. For each pole, we collect 10 images with varying GSV heading, keeping fov and pitch constant. The confidence scores of the detection results obtained from the YOLO model and the end-of-pipeline results showing pole deflection angles for the 100 poles are depicted in figure \ref{fig:results1}. The average pole inclination in our eval set was reported to be 88.83 degrees. Evaluating these results against ground truth is challenging because of the lack of labeled data. However, collecting ground truth data to evaluate our pipeline is the next step for this approach.

\subsection{Depth Estimation}
\begin{figure}[H]
\centering
  \includegraphics[height=3.5cm, width=7.5cm]{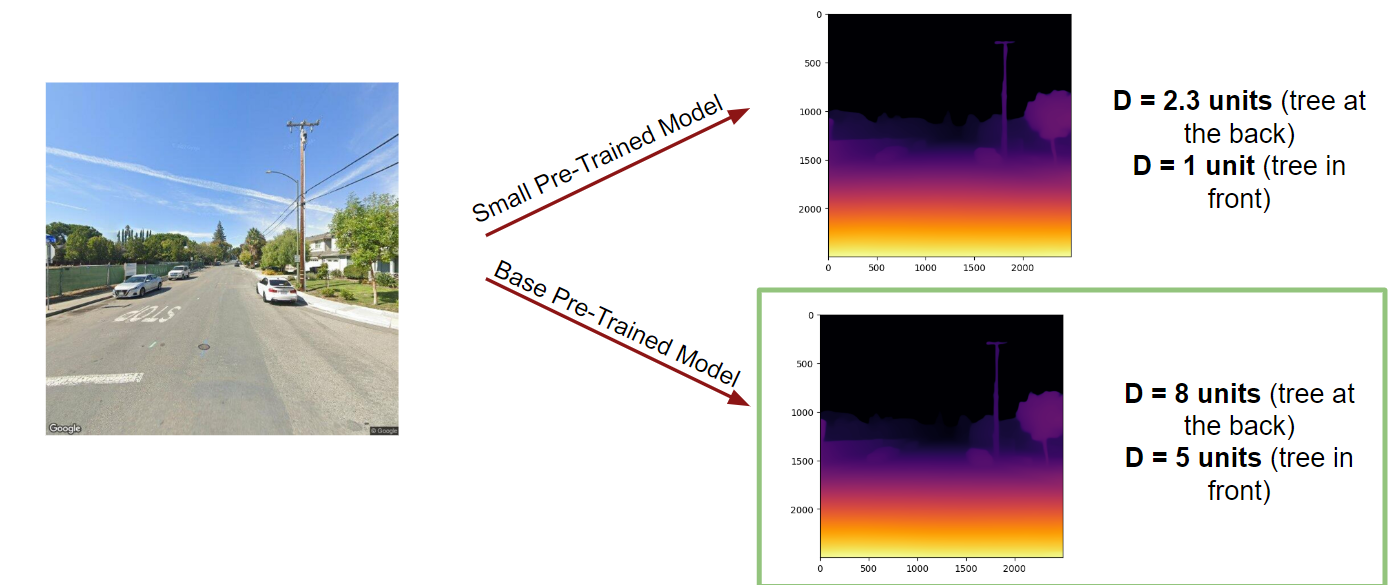}
    \caption{Quantitative Results for Depth Estimation}
    \label{fig:depthres1}
\end{figure}

\begin{figure}[H]
\centering
  \includegraphics[height=4cm, width=7cm]{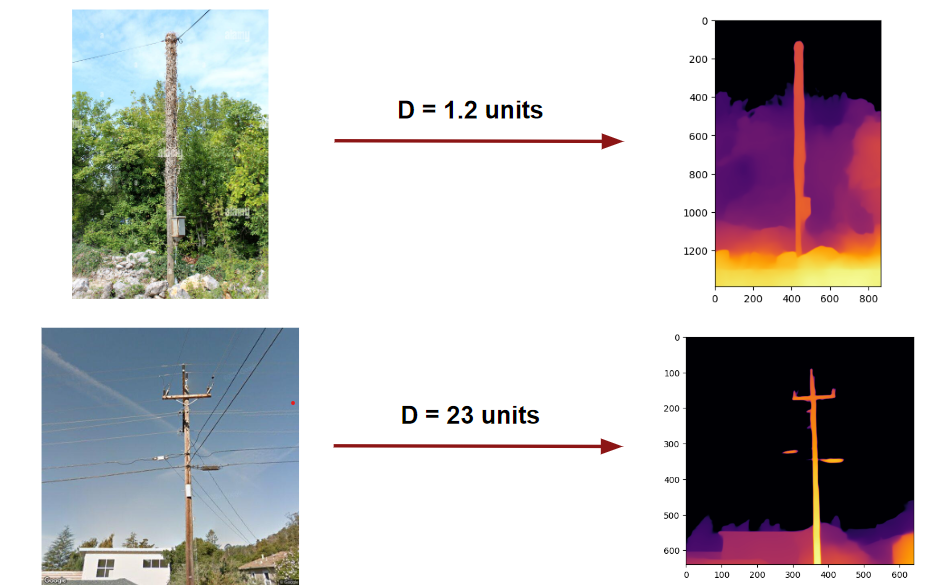}
    \caption{Quantitative Results for Depth Estimation}
    \label{fig:depthres2}
\end{figure}

We obtain results for both "Small" and "Base" pre-trained versions of the Depth Anything Model for selected street-view images. However, we chose the "Base" model for reporting results as it was observed to have performed better qualitatively. We calculate the relative distance as the difference between the absolute values of the depth of the pole and the depth of the surrounding tree. In Figure \ref{fig:depthres1}, we observe the relative depth of the pole to be 8 units from the tree at the back and 5 units from the tree in front. When the pole is heavily surrounded by trees (as in top Figure \ref{fig:depthres2}), the relative depth was observed to be 1.2 units. On the other hand, the relative depth of the pole from the tree was 23 units when the pole was far away from the tree as in the bottom image of Figure \ref{fig:depthres2}. Qualitatively, these results are an underestimate of the actual distance between the pole and trees, however, they prove to be a good starting point to approximate the distance between the pole and trees in order to assess wildfire risk. For the next steps, we propose to compare and correlate the Relative Depth from the actual distance observed on Google Maps to facilitate the generation of heat risk maps based on proximity.

\subsection{3D Point Cloud Reconstruction}
\begin{figure}[H]
\centering
  \includegraphics[height=3.8cm, width=8cm]{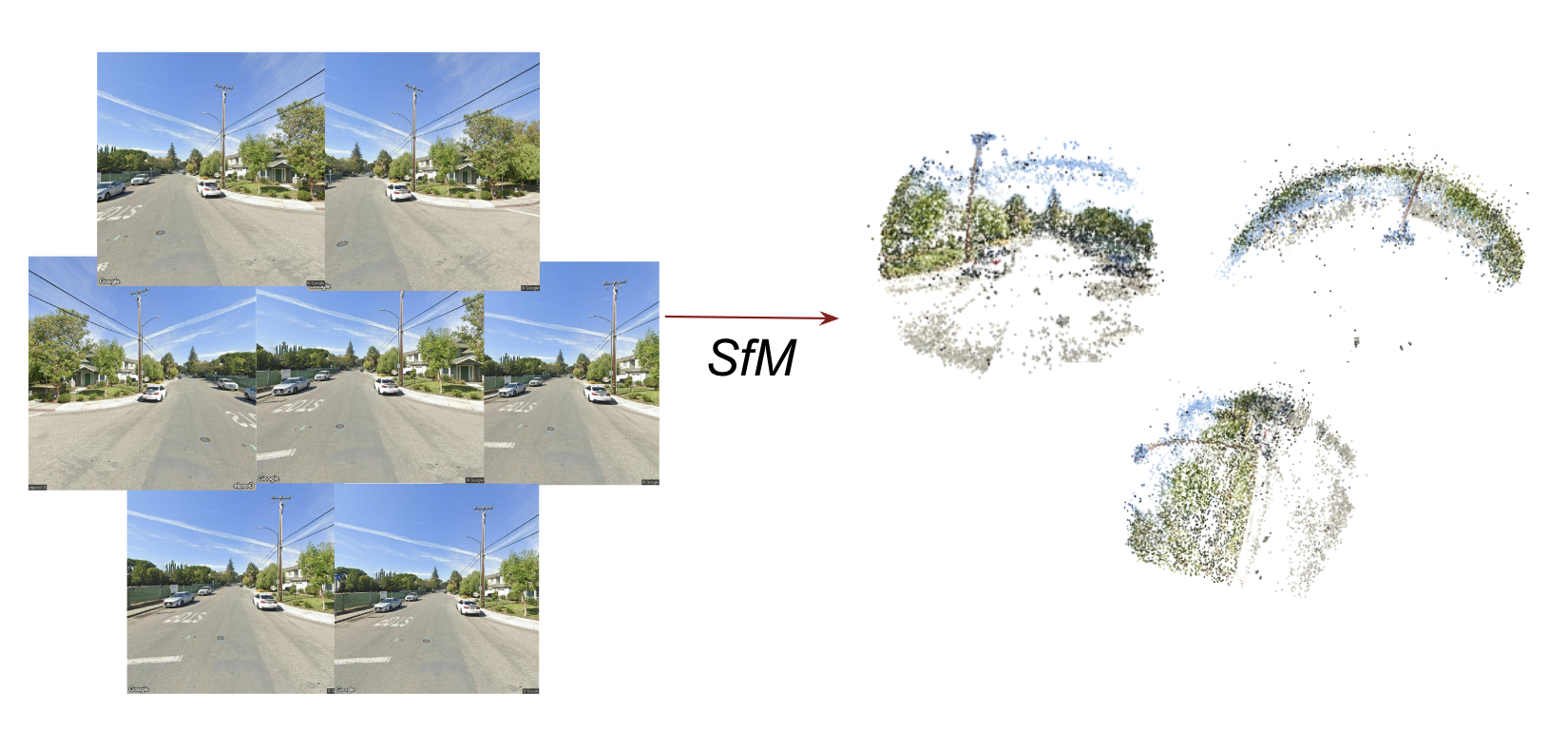}
    \caption{Quantitative Results for 3D Point Cloud Reconstruction of the utility pole and surroundings}
    \label{fig:pcloud_result}
\end{figure}

Quantitative results from the 3D point cloud reconstruction pipeline indicate the utilization of 35 images captured from diverse viewpoints around the utility pole, subsequently processed through the Structure for Motion technique to generate a comprehensive 3D representation. An illustrative example is depicted in the figure \ref{fig:pcloud_result}, albeit with notable limitations. Specifically, the reconstructed point cloud exhibits deficiencies resulting in a distorted view primarily from front-facing perspectives (absence of images capturing the rear aspect of the pole). Moreover, the employed methodology is observed to be computationally demanding, presenting challenges in terms of processing resources and efficiency.

\section{Business Impact}
This research is crucial for the Asset Condition and Facility Management (ACFM) industry, offering a comprehensive approach to wildfire risk assessment and mitigation, thus enhancing the safety and reliability of utility poles. It benefits utility companies by proactively identifying risk (fire and/or topple) areas near utility poles, ensuring worker and resident safety, and informing decision-making processes related to vegetation management, infrastructure upgrades, and emergency response planning. Government agencies overseeing safety standards and environmental regulations also stand to gain valuable insights. Additionally, residents in wildfire-prone areas benefit from enhanced safety and quality of life, while organizations specializing in the ACFM industry can optimize operational efficiencies and improve asset reliability. Overall, implementing these strategies will enhance infrastructure resilience, ensure public safety, mitigate the socio-economic impacts of wildfire events, and will lead to cost savings for utilities.

The infrastructure required for our project includes unmanned aerial vehicles (UAVs) to monitor inaccessible areas, computing resources for continuous updates of infrastructure data, and integration with existing asset management systems. 

\section{Conclusions}
We present a robust framework for assessing the risk associated with utility poles using advanced computer vision techniques. Our approach addresses two key risk factors: the inclination angle of poles and their proximity to vegetation. Leveraging Google Street View images alongside electric utility pole data provided by PG\&E, we explored multiple avenues to quantify these risks.

One aspect of our methodology involved  detecting and localizing poles, followed by utilizing the Hough transform to calculate inclination angles. This angle provides insight into the susceptibility of poles to hazards under high wind loads. Additionally, we employed depth estimation model to gauge the proximity of poles to trees. We also explored 3D point cloud reconstruction for all-encompassing pipeline analysis. In terms of evaluation, we propose project-centric metrics encompassing accuracy, precision, and recall to assess the efficacy of our methods. Furthermore, we introduce impact metrics and cost vulnerability metrics to provide a comprehensive understanding of the potential benefits and risks associated with our framework.

This research holds significant implications for the ACFM industry, offering a proactive approach to mitigate wildfire risks by assessing the safety and resiliency of utility poles. By serving as an early warning system, our framework has the potential to inform critical decision-making processes, ultimately safeguarding infrastructure and communities against the devastating impacts of extreme weather events such as wildfires and wind storms. Additionally, our framework could be utilized to prioritize the undergrounding of high-risk utility poles and power lines as well as aid with public safety power shutoffs (PSPS) events.

\section{Limitations and Future Work}
The project faces some limitations, including potential inaccuracies in distance measurements, variations in vegetation types, occlusions in collected data, and the absence of street view imagery in remote locations, notably where utility poles are surrounded by dense vegetation. Moreover, the quality of images significantly impacts the accuracy of 3D point cloud reconstruction. Addressing these limitations requires high-quality data collection, continuous improvement, and validation processes to minimize disruptions to the ACFM process.

In future work, we aim to incorporate temporal aspects in our risk model by detecting changes in vegetation around poles over time and considering historically reported ignitions near poles. Additionally, generating heat risk maps based on both pole tilt angles and vegetation proximity would enhance the project's effectiveness.

In conclusion, this research study presents a comprehensive and cost-effective approach to evaluating utility pole-induced fire and topple risk using computer vision. Its focus on providing actionable recommendations to prevent wildfires in the ACFM industry underscores its importance and potential impact.

{\small
\bibliographystyle{ieeetr}
\bibliography{egbib}
}

\end{document}